\documentclass{JUSTC}

\usepackage{algorithm,algorithmicx,algpseudocode}%
\usepackage{epsfig}
\usepackage{multirow}
\usepackage{CJKutf8}

\type{Letter}
\title{Deep CLAS: Deep Contextual Listen, Attend and Spell}

\author[1,\Letter]{Mengzhi Wang}

\author[1,2]{Shifu Xiong}

\author[1,2]{Genshun Wan}

\author[2]{Hang Chen}

\author[1]{Jianqing Gao}

\author[2]{Lirong Dai}

\email{mzwang@ustc.edu}

\runningtitle{Deep Contextual Listen, Attend and Spell}
\runningauthor{Wang \etal}

\address{iFLYTEK Research, iFLYTEK Co. Ltd., Hefei 230088, China}

\address{National Engineering Research Center of Speech and Language Information Processing, University of Science and Technology of China, Hefei 230027, China}

\abstract{
  Contextual-LAS (CLAS) has been shown effective in improving Automatic Speech Recognition (ASR) of rare words. It relies on phrase-level contextual modeling and attention-based relevance scoring without explicit contextual constraint which lead to insufficient use of contextual information. In this work, we propose deep CLAS to deeply utilize contextual information. We introduce bias loss forcing model to focus on contextual information. The query of bias attention is also enriched to improve the accuracy of the bias attention score. To get fine-grained contextual information, we replace phrase-level encoding with character-level encoding and encode contextual information with conformer. Furthermore, the bias attention score is directly utilized to correct the model's output probability distribution. Additionally, a prefix tree is employed to prevent interference from irrelevant information. Experiments using the public AISHELL-1. Compared to CLAS baselines, deep CLAS obtains a 65.78\% relative recall and a 53.49\% relative F1-score increase in the named entity recognition scene.
}

\keywords{Speech Recognition; CLAS; Deep Context}

\begin{document}
\begin{CJK}{UTF8}{gkai}

\maketitle 

\section{Introduction}

Speech recognition is one of the most popular human-computer interaction methods at present. Thanks to the development of deep learning technology and the accumulation of data in recent years\textsuperscript{\cite{gulati20_interspeech,han20_interspeech,radford2022robustspeechrecognitionlargescale,peng24b_interspeech}}, the accuracy of machine speech recognition in some scenes has exceeded that of humans\textsuperscript{\cite{xiong2016achieving}}. However, the data-driven deep learning model suffers long-tail distribution problems\textsuperscript{\cite{czarnowska2019don,longtail}}. Rare words, such as entity names, are hard to recognize.

There are two mainstream methods to improve the recognition rate of rare words in the traditional ASR system. One is adding more training text about rare words of the language model\textsuperscript{\cite{mcgraw2016personalized,8682490}}. Another way is to construct a prefix tree\textsuperscript{\cite{jung2022spell}} to increase the corresponding posterior probability during decoding.

Compared with the traditional model, the end-to-end model significantly improves speech recognition accuracy in general scenes\textsuperscript{\cite{vielzeuf2021e2e}}. However, using the above two methods in the end-to-end model is challenging because the end-to-end model has no pronunciation dictionary and uses a word or sub-word modeling\textsuperscript{\cite{zeineldeen2020systematic}} directly. As a result, the model may assign a higher probability to homophonic words, making the contextualization very difficult to take effect. Therefore, the end-to-end model usually introduces contextual information into the model and optimizes jointly to improve the contextualization effect. Accordingly, the Contextual Listen-Attend-Spell (CLAS)\textsuperscript{\cite{pundak2018deep,8682738,chen19l_interspeech}} model was proposed to encode a set of contextual phrases first. Then the state of the decoder is used to extract relevant contextual information through the attention mechanism. Finally, the context is passed to decoder layers to obtain the output probability. A particular $\langle/bias\rangle$ symbol is inserted at the corresponding positions of the reference to make the model pay attention to the contextual information.

We make a deep analysis of the CLAS model and have some interesting findings. First, the correct recognition of $\langle/bias\rangle$ does not mean the correct recognition of bias words. The model tends to output $\langle/bias\rangle$ directly when the decoding history matches a bias word and ignores contextual information before output $\langle/bias\rangle$. Second, CLAS uses the last decoder state as the query of bias attention, which is missing the last recognized word information. However, the model should pay more attention to the bias words that contain the last recognized word. Furthermore, the current acoustic context is vital for bias attention because the acoustic context contains the pronunciation information of the current word, which is not contained in the last decoder state. Third, CLAS uses coarse-grained contextual information encoding. Each bias word has only one embedding. The single embedding may not be able to encode the information of all characters if the bias word is long. Finally, CLAS feeds contextual information into the decoder, but the model may not use it at all. Therefore, CLAS does not make effective use of contextual information.

Building on the above discussion, we proposed a novel deep CLAS model to improve the contextualization performance of the CLAS model. First, we introduce the bias loss forcing model to focus on contextual information. Second, we use the last decoder state, the last decoded word and the current acoustic context together as the query of bias attention to improve the accuracy of bias attention score. Third, we replace phrase-level encoding with character-level encoding to obtain fine-grained contextual information. Fourth, we explicitly use the bias attention score to utilize contextual information more effectively. Finally, we use the prefix tree to reduce the interference of irrelevant bias words, resulting in a better effect in the long bias word recognition scenario.

\section{Related work}
\begin{figure}[h!t]
\begin{minipage}[b]{.5\linewidth}
  \centering
  \centerline{\epsfig{figure=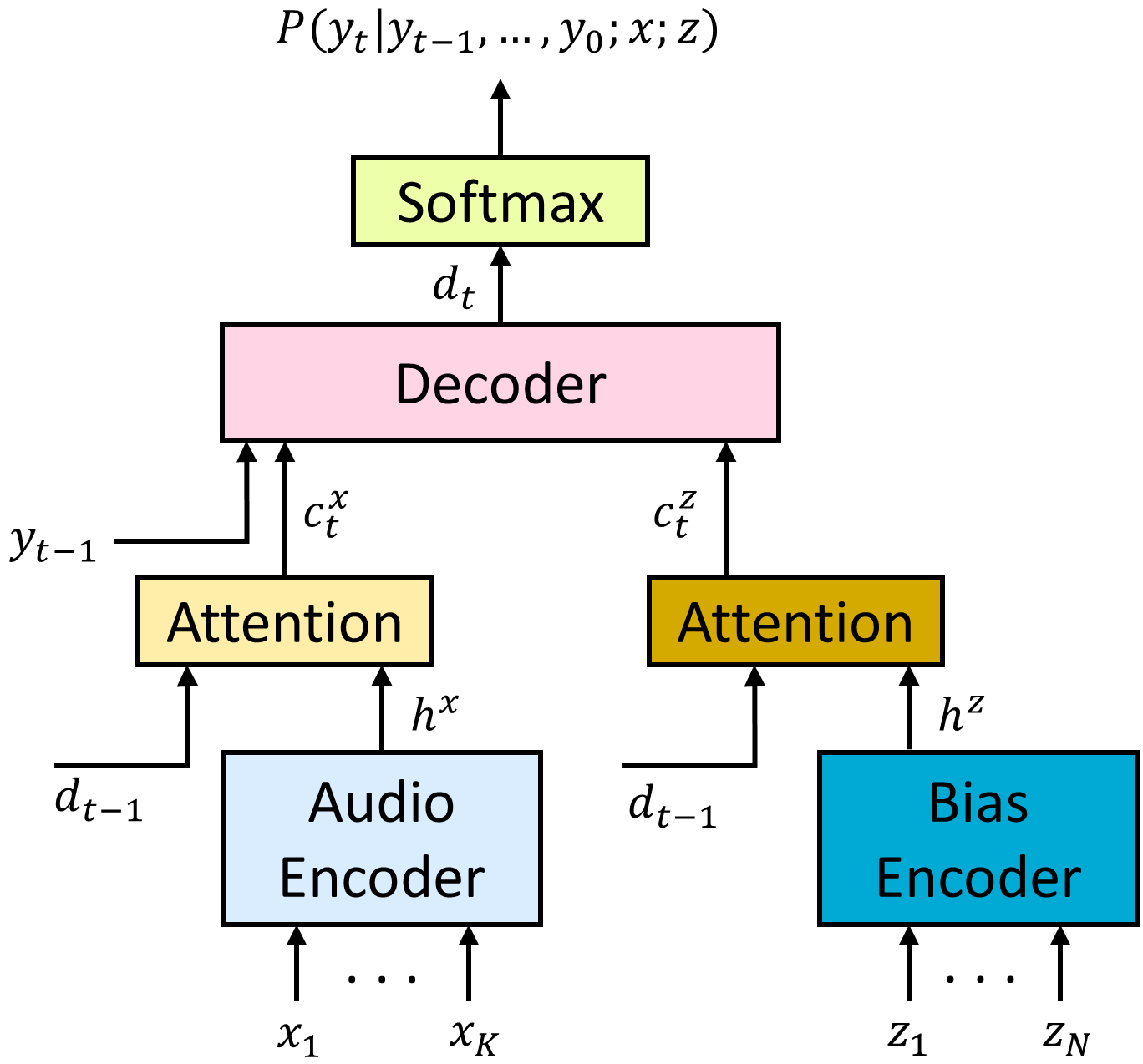,width=5.0cm}}
  \centerline{(a) CLAS Model}\medskip
\end{minipage}
\hfill
\begin{minipage}[b]{0.5\linewidth}
  \centering
  \centerline{\epsfig{figure=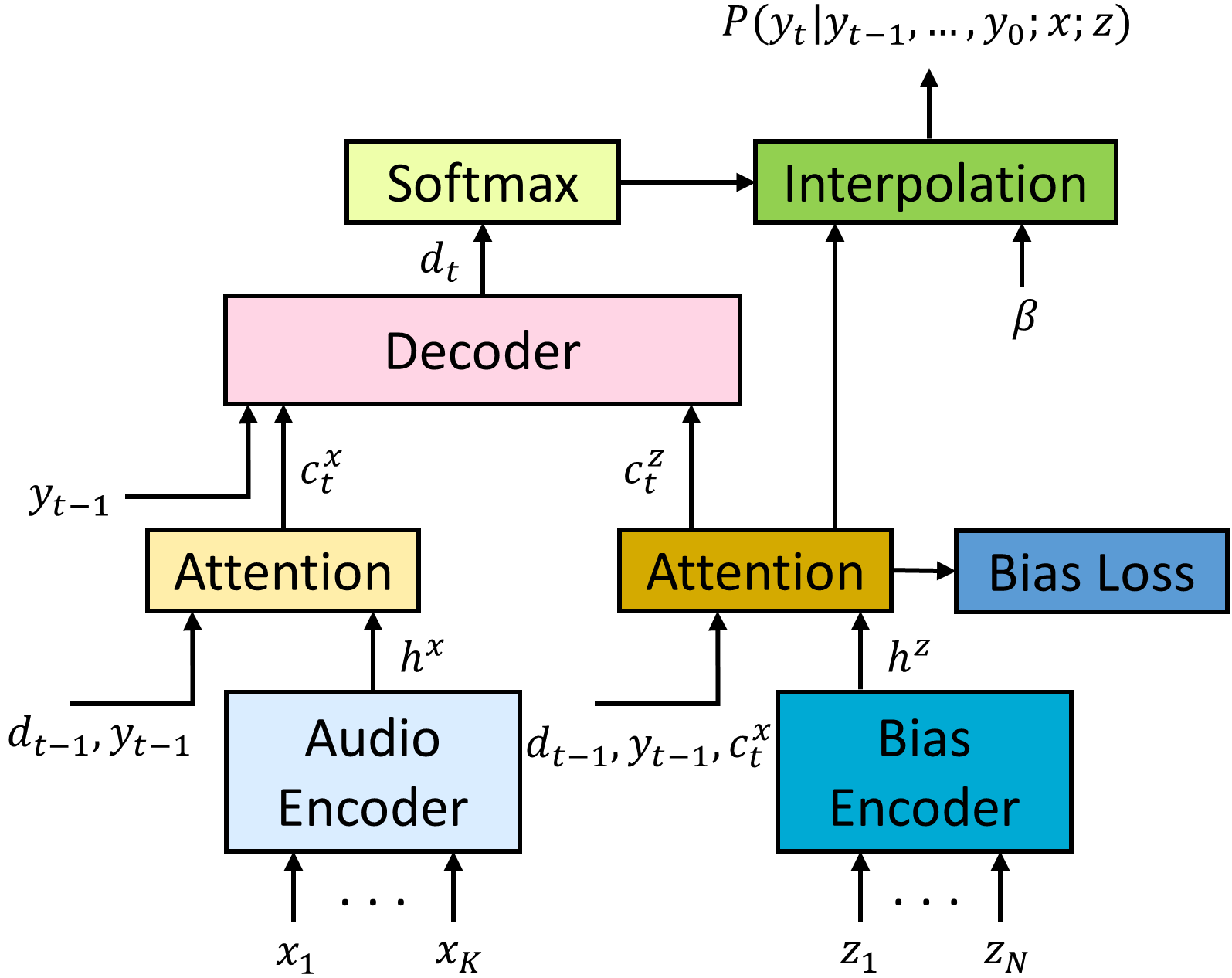,width=5.0cm}}
  \centerline{(b) Deep CLAS Model}\medskip
\end{minipage}
\caption{The framework of CLAS model and deep CLAS model.}
\label{fig:CLAS}
\end{figure}

The CLAS model derives from the Listen, Attend and Spell (LAS) model\textsuperscript{\cite{chan2015listen}} but increases the use of contextual information. This model has been proven to be an effective way for contextual biasing. Fig. \ref{fig:CLAS}(a) illustrates the overall structure of the CLAS model.

CLAS model contains two encoders, one decoder and two attention networks. One encoder is an audio encoder and the other is a bias encoder. The audio encoder encodes speech frames $x =(x_1,...,x_K)$ into high-level acoustic features $h^x$. And bias encoder encodes bias phrases $z = (z_1,...,z_N)$ into high-level contextual features $h^z$. The bias encoder is a multilayer long short-term memory network (LSTM)\textsuperscript{\cite{10.1162/neco.1997.9.8.1735}}. The last state of the LSTM is considered as the embedding of the entire bias phrase. Since the bias phrases may not be relevant to the current utterance, CLAS includes an additional learnable vector $h_{nb}^{z}$, corresponding to the ‘no-bias’ option. Similar to LAS, CLAS generates acoustic context vector $c^x$ using the previous decoder state $d_{t-1}$ and the output of the audio encoder $h^x$ through the audio attention network.
\begin{equation}\label{eq1} 
  u_{it}^{x}=v^{x}tanh(W_{h}^{x} h_{i}^{x}+W_{d}^{x} d_{t-1}+b_{a}^{x})
\end{equation}
\begin{equation}\label{eq2} 
  \alpha_{t}^{x}=softmax(u_{t}^{x})
\end{equation}
\begin{equation}\label{eq3} 
  c_{t}^{x}=\sum_{i=0}^{K}{\alpha_{it}^{x} h_{i}^{x}}
\end{equation}

Then generate bias context vector $c^z$ using the previous decoder state $d_{t-1}$ and the output of bias encoder $h^z$ through bias attention network.
\begin{equation}\label{eq4} 
  u_{it}^{z}=v^{z}tanh(W_{h}^{z} h_{i}^{z}+W_{d}^{z} d_{t-1}+b_{a}^{z})
\end{equation}
\begin{equation}\label{eq5} 
  \alpha_{t}^{z}=softmax(u_{t}^{z})
\end{equation}
\begin{equation}\label{eq6} 
  c_{t}^{z}=\sum_{i=0}^{N}{\alpha_{it}^{z} h_{i}^{z}}
\end{equation}

After obtaining $c^x$ and $c^z$, they are spliced and fed into the decoder to get the decoder state $d_t$ and output probabilities over all modeling units.
\begin{equation}\label{eq7} 
  d_t=RNN(d_{t-1},y_{t-1},c_t^x,c_t^z)
\end{equation}
\begin{equation}\label{eq8} 
  P(y_t | h^x,y<t,z)=softmax(W_s d_t+b_s)
\end{equation}
\begin{equation}\label{eq9} 
  L_{CLAS}=-logP(y | x,z)
\end{equation}

During training, the bias phrase list is randomly created from the reference transcripts associated with the utterances in the training batch. To exercise the ‘no-bias’ option, not all reference transcripts extract bias words. The proportion of extracting bias words is $P_{keep}$. $k$ word n-grams are randomly selected from each kept reference, where $k$ is picked uniformly from $[1,N_{phrases}]$ and $n$ is picked uniformly from $[1,N_{order}]$. $P_{keep}$, $N_{phrases}$ and $N_{order}$ are hyperparameters of the training process. A special $\langle/bias\rangle$ symbol is inserted where bias phrases appear in reference transcripts. The purpose of $\langle/bias\rangle$ is to introduce a training error which can be corrected only by considering the correct bias phrase.

\section{Materials and methods}

In this section, we will introduce the proposed deep CLAS model. Fig. \ref{fig:CLAS}(b) illustrates the overall structure of the deep CLAS model.

\subsection{Bias loss}
The CLAS inserts $\langle/bias\rangle$ into the reference transcripts to make the model focuses on contextual information. However, the model can ignore contextual information and output $\langle/bias\rangle$ when the historical output is the same as a bias word. Besides, the model does not know when to start paying attention to contextual information. Therefore, we used the bias loss\textsuperscript{\cite{zhang2022end}}, a cross-entropy loss of the bias attention coefficient. The bias loss makes the model focuses on the corresponding bias words in the bias list when decoding bias words. Otherwise, it focuses on the ‘no-bias’. The bias loss is defined as:
\begin{equation}\label{biasloss} 
  L_{bias}=-\sum_t\sum_{k}y_{tk}log(\alpha_{tk})
\end{equation}
Where $\alpha_{tk}$ is the bias attention score of the kth word of the bias list at the decoding step $t$, and $y_t$ is a one-hot vector. If the t-th word of reference is part of the kth bias word, $y_{tk}$ is 1; otherwise, it is 0. If there is no bias word match, $y_{t0}$ is 1 which indicates attending ‘no-bias’. The bias loss is combined with the standard CTC/attention loss as the optimization object: 
\begin{equation}\label{loss} 
  L= (1 - \lambda)L_{att}+\lambda L_{ctc}+L_{bias}
\end{equation}

\subsection{The query of bias attention}
The CLAS uses $d_{t-1}$ as the query of the bias attention. Fig. \ref{fig:CLAS}(a) shows $d_{t-1}$ does not contain the last recognized word information $y_{t-1}$. However, the last recognized word is essential for bias attention. A bias word can have a higher attention score if it contains $y_{t-1}$ or a lower attention score if not. Besides, $d_{t-1}$ lacks of $c_{t}^{x}$ which is acoustic contextual information at the current time. A bias word should have a higher attention score if it matches acoustic information at the current time rather than historical time. Thus, we modify Eq. \ref{eq4} to:
\begin{equation}\label{biasatt} 
  u_{it}^{z}=v^{z}tanh(W_{h}^{z} h_{i}^{z}+W_{d}^{z} [d_{t-1}:y_{t-1}:c_{t}^{x}]+b_{a}^{z})
\end{equation}
$[d_{t-1}:y_{t-1}:c_{t}^{x}]$ indicates splicing $d_{t-1}$, $y_{t-1}$ and $c_{t}^{x}$. In this way, the model has more information when scoring bias words, so bias words are scored more accurately.

\subsection{The granularity of bias information}
The CLAS model utilizes phrase-level embeddings to represent bias words, with each bias word being represented by a fixed-size vector. However, there is a potential problem of information loss when the bias word is long. Furthermore, phrase-level embeddings do not correspond to the output modeling unit, which can be either sub-words or characters. In this case, the model doesn't know which character in the bias word to output when using phrase-level embeddings. To address these issues, we use the output of all time steps of the LSTM instead of just the last state, to obtain fine-grained contextual information. Additionally, we integrate the BLSTM and conformer architecture to replace the LSTM. This change results in better embeddings.

\subsection{Probability fusion}
The CLAS model feeds contextual information into the decoder, making the model choose whether to use the contextual information, which is an implicit way to use contextual information and may lead to performance degradation. To use contextual information explicitly, we modify the probability distribution of the model output according to the bias attention score directly. This is done by scaling the bias attention score and the probability distribution of the model output and then combining them to obtain a final probability distribution:
\begin{equation}\label{fusion1} 
  P_{b}^{1…z}=\beta P_{b}^{1…z}
\end{equation}
\begin{equation}\label{fusion2}
  P_{m}=(1-\beta(1-P_{b}^{0}))P_m
\end{equation}
In the formula, $P_b$ denotes the probability distribution of bias attention and $P_m$ denotes the probability distribution of model output. $P_{b}^{0}$ is the score of ‘no-bias’ and $P_{b}^{1…z}$ is the score of bias words. $\beta$ is the bias coefficient to control the bias degree. The larger the $\beta$ is, the easier it is for the model to recognize bias words, but more false triggers may occur. We first multiply $P_{b}^{1…z}$ by $\beta$, so the sum of $P_{b}^{1…z}$ is $\beta(1-P_{b}^{0})$. The remaining score is $1-\beta(1-P_{b}^{0})$, so we multiply $P_m$ by $1-\beta(1-P_{b}^{0})$. Then $P_m$ plus $P_{b}^{1…z}$ equals one. There are two ways to combine $P_m$ and $P_{b}^{1…z}$. One way is to concatenate $P_{b}^{1…z}$ behind $P_m$ to form a larger vocabulary just like pointer generator\textsuperscript{\cite{see2017get}}. In fine-grained contextual encoding, since a character may appear in multiple bias words, we can merge the bias scores of the same character from different bias words to make the bias scores more concentrated. Another way is to merge the scores of the same characters in $P_m$ and $P_{b}^{1…z}$, which we call interpolation. In this case, the vocabulary size does not change.

\subsection{Prefix tree}
\begin{figure}[h!t]
\centering
\includegraphics[width=.5\linewidth]{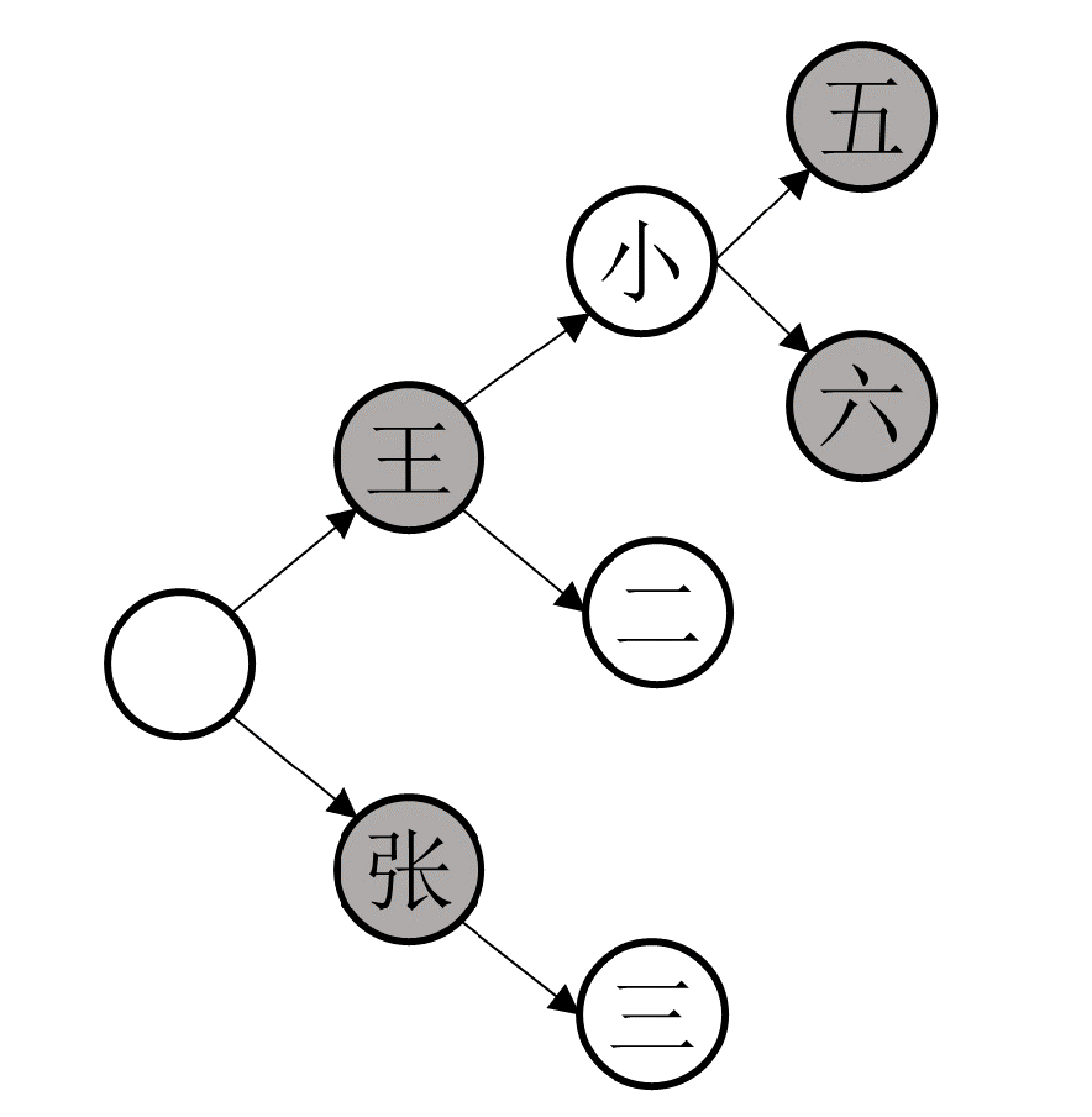}
\caption{Prefix tree when a list of bias words consists of (“张三”,“王二”, “王小五” , “王小六”) in character level.}
\label{fig:tree}
\end{figure}

The prefix tree\textsuperscript{\cite{sun2021tree,9383560,le21_interspeech}} is widely used in the traditional bias method. Building a prefix tree using bias words, which is traversed during decoding, can prevent the excitation of bias words whose prefix does not match the decoding history. Because only active tree nodes in the prefix tree will be excited. Fig. \ref{fig:tree} illustrates a prefix tree when the decoding history is “王小”. The active tree nodes are represented by circles with gray background. The CLAS uses the last decoder state to score bias words. Theoretically, it can achieve the same effect as the prefix tree. However, neural networks tend to overfitting and the effect drops sharply in mismatched scenes. The longest bias word during training is only four characters, which may reduce the effect of bias words with more than four characters. Therefore, we conducted a long bias word experiment and used the prefix tree to solve the mismatch problem.

\section{EXPERIMENTS}

\subsection{Datasets}
We conduct experiments on AISHELL-1\textsuperscript{\cite{aishell}}, which is a benchmark dataset for mandarin ASR. The AISHELL-1 dataset contains 178 hours of data collected from 400 people. The test set of AISHELL-1 does not contain bias words. To make the results reproducible, we use the named entities of the AISHELL-NER\textsuperscript{\cite{chen2022aishellnernamedentityrecognition}} as bias words. There are 3393 named entities in the test set. We retain named entities containing two to four characters and de-duplicate them to get 891 named entities as the bias words list.

\subsection{Experimental Setup}
Our model is based on the joint CTC-attention structure. CTC weight is set to 0.3. The speech features are 80-dimensional log-mel filterbank (FBank) computed on 32ms window with 10ms shift. We apply SpecAugment\textsuperscript{\cite{park2019specaugment}} with mask parameters $(F=10,T=50,m_F=m_T=2)$ and speed perturbation\textsuperscript{\cite{ko15_interspeech}} with factors of 0.9, 1.0 and 1.1 to augment training data. The encoder is composed of a subsampling module and 12-layer conformer blocks. The subsampling module contains two Conv2d with kernel size 3*3 and stride 2*2, resulting in 40ms framerate. For conformer blocks, the attention dimension is 256 with 4 attention heads, the dimension in the feed-forward network sublayer is 2048, and the kernel size of the conformer convolution module is 15. The decoder is a layer of LSTMP with a hidden size of 1024 and a projection size of 512. The bias encoder is a layer of (Bi-directional) LSTMP with a hidden size of 512 and a projection size of 256 or two layers of conformer with same configuration as encoder. The encoder attention is computed over 128-dimensional location sensitive attention\textsuperscript{\cite{chorowski2015attention}}. The bias attention is computed over 512-dimensional additive attention. Word embedding size is 512. We set attention dropout to 0.1 and label smooth\textsuperscript{\cite{7780677}} to 0.1. We apply the Adam\textsuperscript{\cite{kingma2017adammethodstochasticoptimization}} optimizer with a gradient clipping value of 5.0 and warmup\textsuperscript{\cite{vaswani2023attentionneed}} learning rate scheduler with 25,000 warmup steps and peak learning rate 0.0016. We train the models for 180 epochs and average the last 30 checkpoints to obtain the final model. Our model is implemented using Pytorch and was trained on one A40 GPU with a batch size 85 and gradient accumulation 3.  So the actual batch size is 255. During training, we set $P_{keep}=0.5, N_{phrases}=1, N_{order}=4$ for consistency with the original CLAS.

To enable the model to obtain the pronunciation information of bias words, we add a phoneme encoder to encode syllable of characters in bias words. Phoneme embeddings and character embeddings are concatenated and reduced in dimension with a linear layer to obtain final embeddings. We use a python package called pypinyin to convert Chinese characters into syllables. Decoding was performed using the beam search decoding algorithm with beam 10 without using any external language model.

We utilize recall, precision and F1-score as our measurement:
\begin{equation}\label{R} 
  R=\frac{n}{N_t}
\end{equation}
\begin{equation}\label{P} 
  P=\frac{n}{N_r}
\end{equation}
\begin{equation}\label{F1} 
  F1=\frac{2*P*R}{P+R}
\end{equation}
Where $R$ is the recall, $P$ is the precision, and $n$ is the number of bias words correctly identified in the recognition result. $N_r$ is the number of all bias words in the recognition result and $N_t$ is the number of all bias words in the reference transcript. F1-score is the harmonic mean of recall and precision. The higher the F1-score, the better. We also calculate character error rate (CER) to measure the overall recognition performance:
\begin{equation}\label{CER} 
  CER=\frac{S+D+I}{N}
\end{equation}
Where $S$ is the number of substitutions, $D$ is the number of deletions, $I$ is the number of insertions, and $N$ is the total number of characters in the reference text. The lower the CER, the better.

\section{Results and Discussion}

In this section, we give the overall results in Table \ref{tab:overall}, and then analyze the methods proposed in this paper.

\begin{table}
\centering
\caption{Results of using all named entities as the bias list. The Y/N in the bias column represents whether bias phrases are present. And the number in the bias column represents the bias coefficient.}
\label{tab:overall}
\begin{tabular}{lccccc}
   \toprule
   Methods & bias & CER & Recall & Precision & F1 \\
   \midrule
   \multirow{2}{*} {E0: CLAS} & N & 4.91 & 79.35 & 97.88 & 87.65 \\
   & Y & 4.92 & 79.25 & 97.80 & 87.55 \\
   \midrule
   \multirow{2}{*} {E1: E0+bias loss} & N & 5.11 & 78.46 & 98.20 & 87.22 \\
   & Y & 5.06 & 79.59 & 98.10 & 87.88 \\
   \midrule
   \multirow{2}{*} {E2: E1+query $y_{t-1}$} & N & 5.04 & 79.04 & 98.04 & 87.52 \\
   & Y & 4.99 & 79.86 & 97.94 & 87.98 \\
   \midrule
   \multirow{2}{*} {E3: E2+query $c_{t}^{x}$} & N & 4.95 & 79.25 & 98.30 & 87.75 \\
   & Y & 4.83 & 82.16 & 97.87 & 89.33 \\
   \midrule
   \multirow{2}{*} {E4: E3+fine-grained} & N & 4.89 & 79.59 & 97.60 & 87.68 \\
   & Y & 4.80 & 82.85 & 98.01 & 89.79 \\
   \midrule
   \multirow{2}{*} {E5: E4+BLSTM} & N & 4.91 & 79.90 & 97.94 & 88.0 \\
   & Y & 4.65 & 85.11 & 97.72 & 90.98 \\
   \midrule
   \multirow{2}{*} {E6: E4+conformer} & N & 4.88 & 79.49 & 98.05 & 87.80 \\
   & Y & 4.60 & 86.86 & 97.80 & 92.01 \\
   \midrule
   \multirow{3}{*} {E7: E6+pointer generator} & 0.1 & 4.49 & 91.66 & 96.01 & 93.79 \\
   & 0.2 & 4.52 & 94.31 & 93.60 & 93.95 \\
   & 0.3 & 4.64 & 95.95 & 90.46 & 93.12 \\
   \midrule
   \multirow{3}{*} {E8: E7+merge probability} & 0.1 & 4.49 & 91.70 & 95.98 & 93.79 \\
   & 0.2 & 4.52 & 94.37 & 93.48 & 93.92 \\
   & 0.3 & 4.65 & 96.09 & 90.27 & 93.09 \\
   \midrule
   \multirow{3}{*} {E9: E6+interpolation} & 0.1 & 4.47 & 92.90 & 95.55 & 94.21 \\
   & 0.2 & 4.55 & 94.72 & 92.84 & 93.77 \\
   & 0.3 & 4.67 & 95.95 & 89.96 & 92.86 \\
   \bottomrule
\end{tabular}
\end{table}

\begin{table}[h!t]
\centering
\caption{Results of using corresponding named entities as the bias list.}
\label{tab:base}
\begin{tabular}{lccccc}
   \toprule
   Methods & bias & CER & Recall & Precision & F1 \\
   \midrule
   \multirow{2}{*} {CLAS} & N & 4.91 & 78.50 & 100 & 87.97 \\
   & Y & 4.90 & 78.50 & 100 & 87.97 \\
   \bottomrule
\end{tabular}
\end{table}

\subsection{Bias loss}
\begin{figure}[h!t]
\begin{minipage}[b]{0.5\linewidth}
  \centering
  \centerline{\epsfig{figure=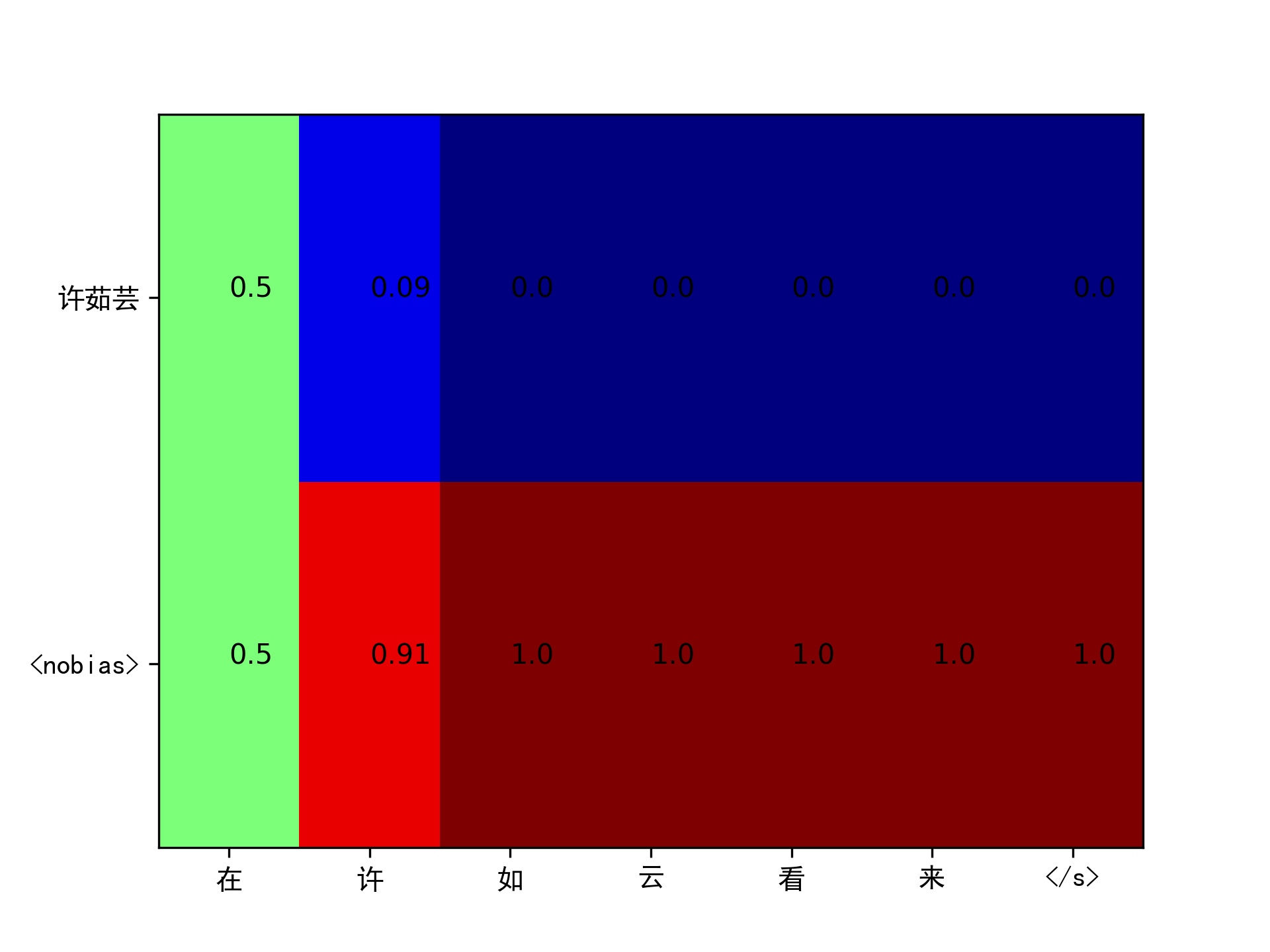,width=6.0cm}}
  \centerline{(a) E0}\medskip
\end{minipage}
\hfill
\begin{minipage}[b]{0.5\linewidth}
  \centering
  \centerline{\epsfig{figure=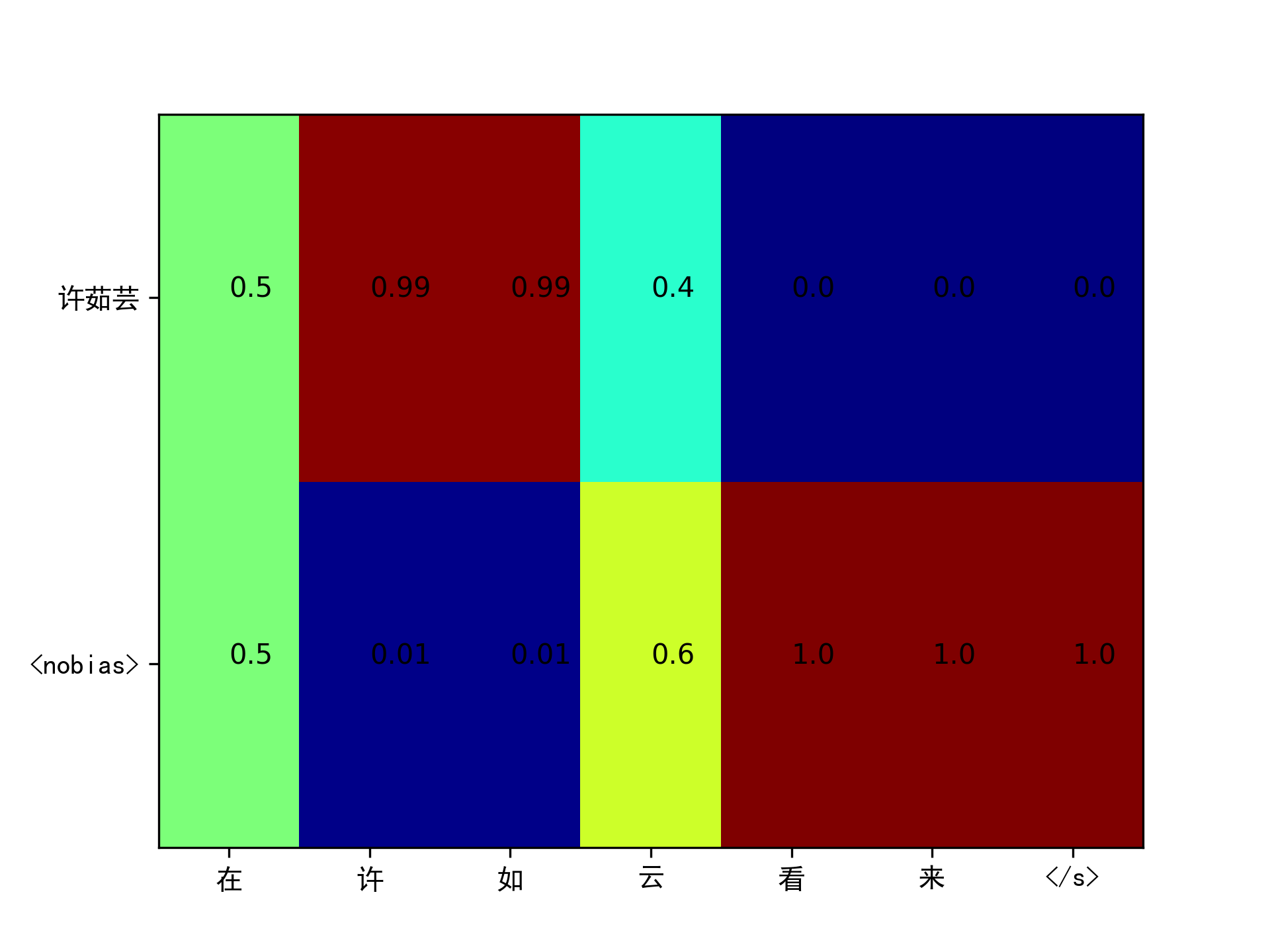,width=6.0cm}}
  \centerline{(b) E1}\medskip
\end{minipage}

\begin{minipage}[b]{0.5\linewidth}
  \centering
  \centerline{\epsfig{figure=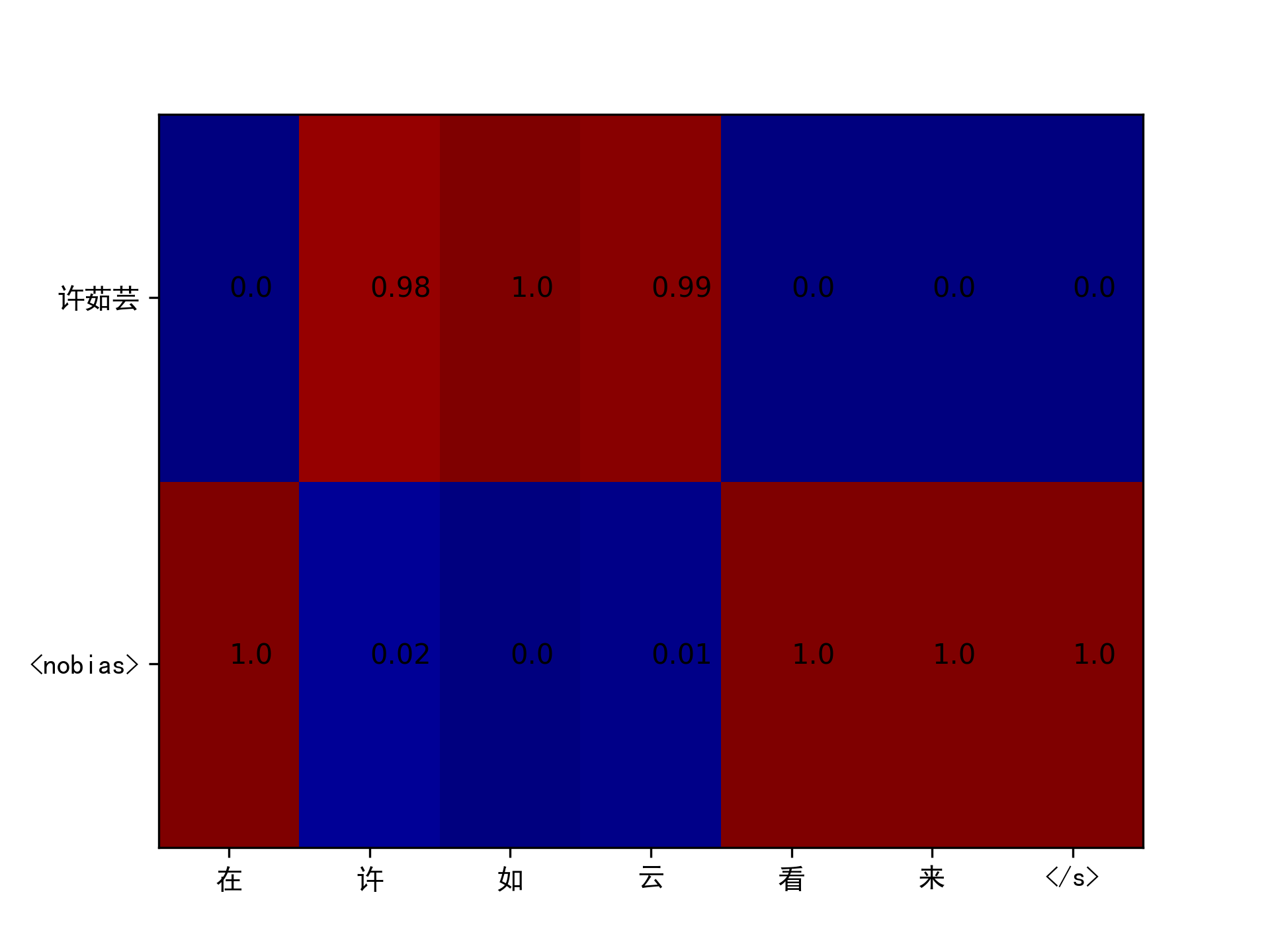,width=6.0cm}}
  \centerline{(c) E3}\medskip
\end{minipage}
\hfill
\begin{minipage}[b]{0.5\linewidth}
  \centering
  \centerline{\epsfig{figure=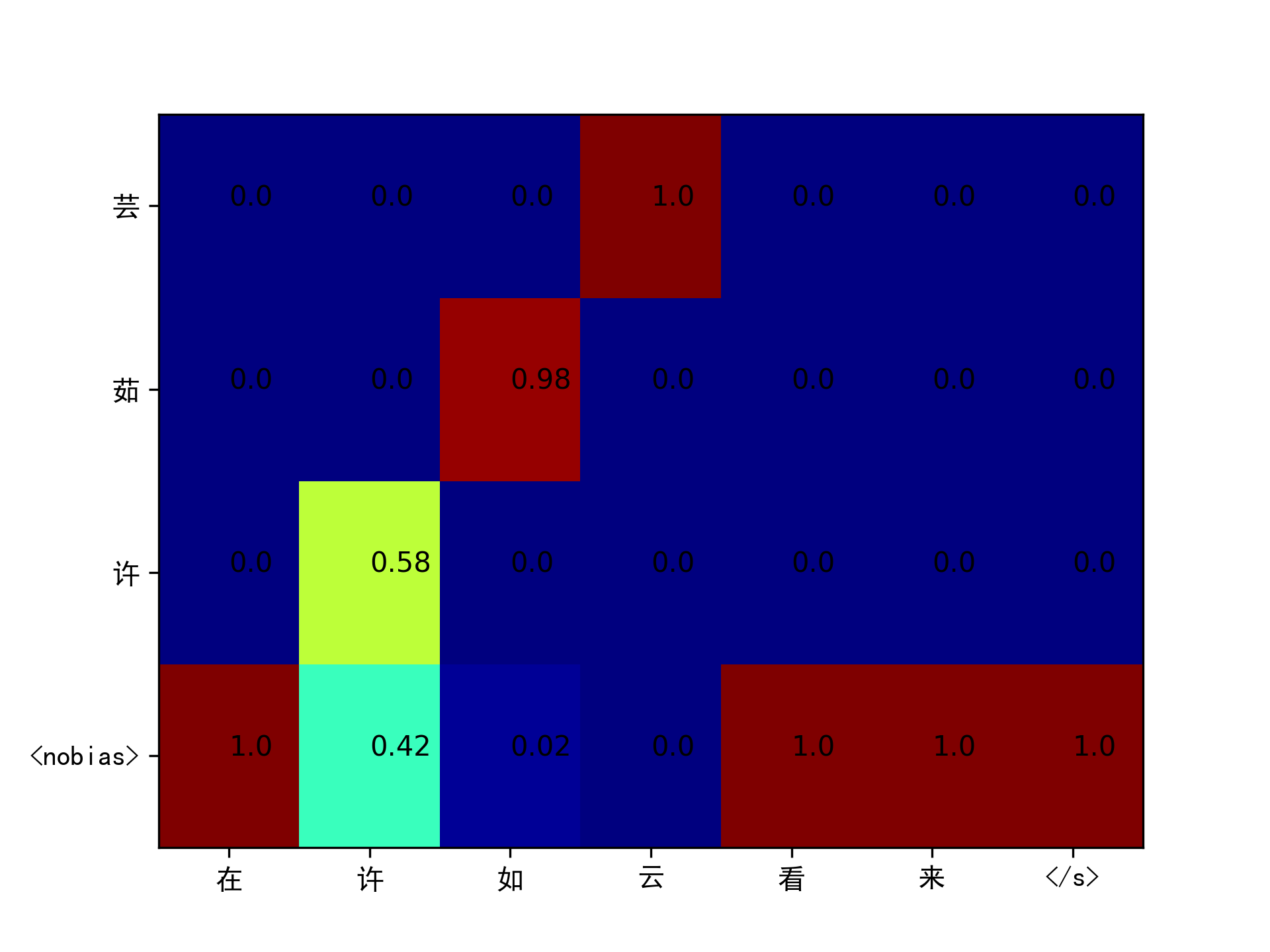,width=6.0cm}}
  \centerline{(d) E4}\medskip
\end{minipage}

\begin{minipage}[b]{0.5\linewidth}
  \centering
  \centerline{\epsfig{figure=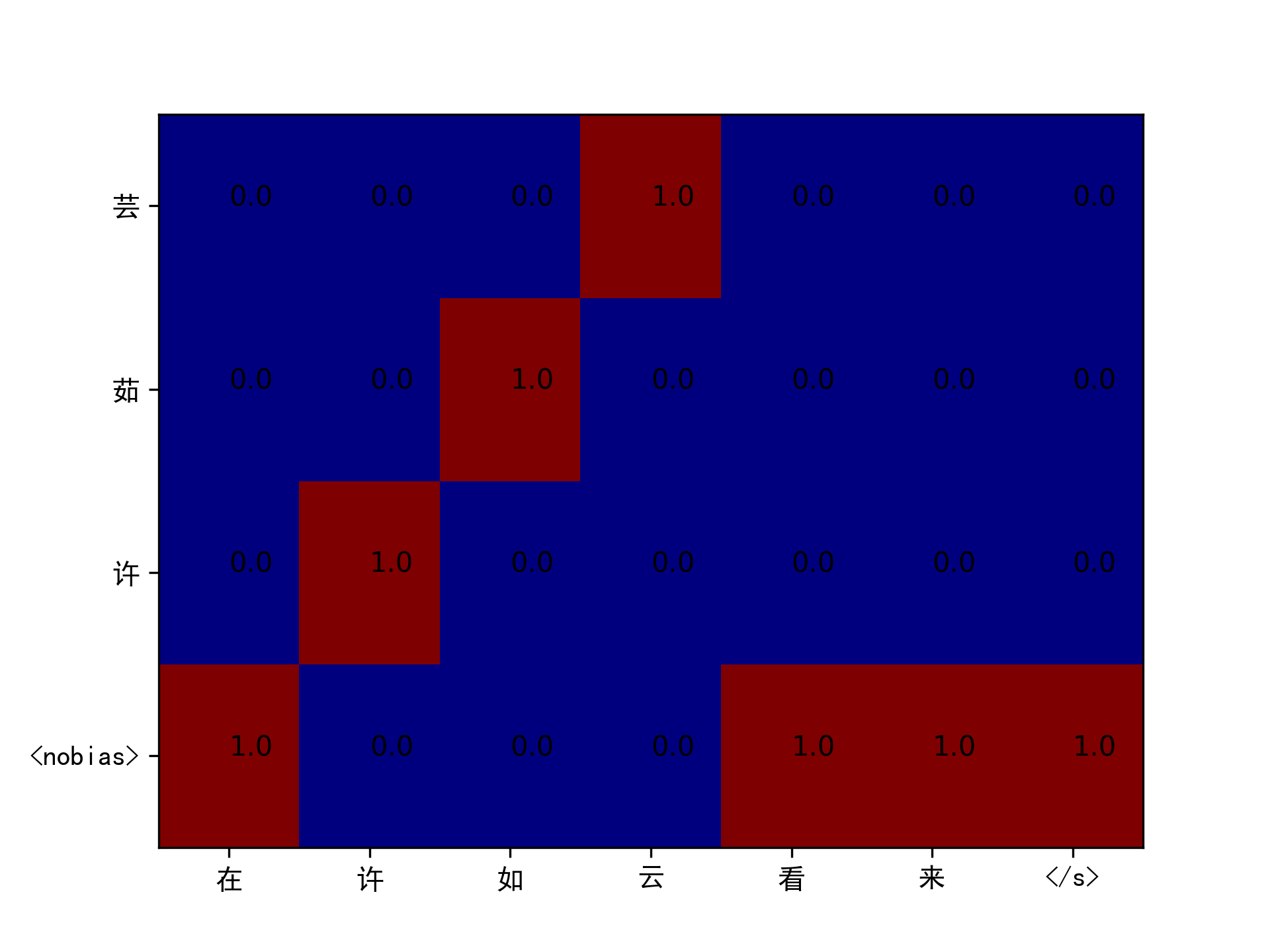,width=6.0cm}}
  \centerline{(e) E6}\medskip
\end{minipage}
\hfill
\begin{minipage}[b]{0.5\linewidth}
  \centering
  \centerline{\epsfig{figure=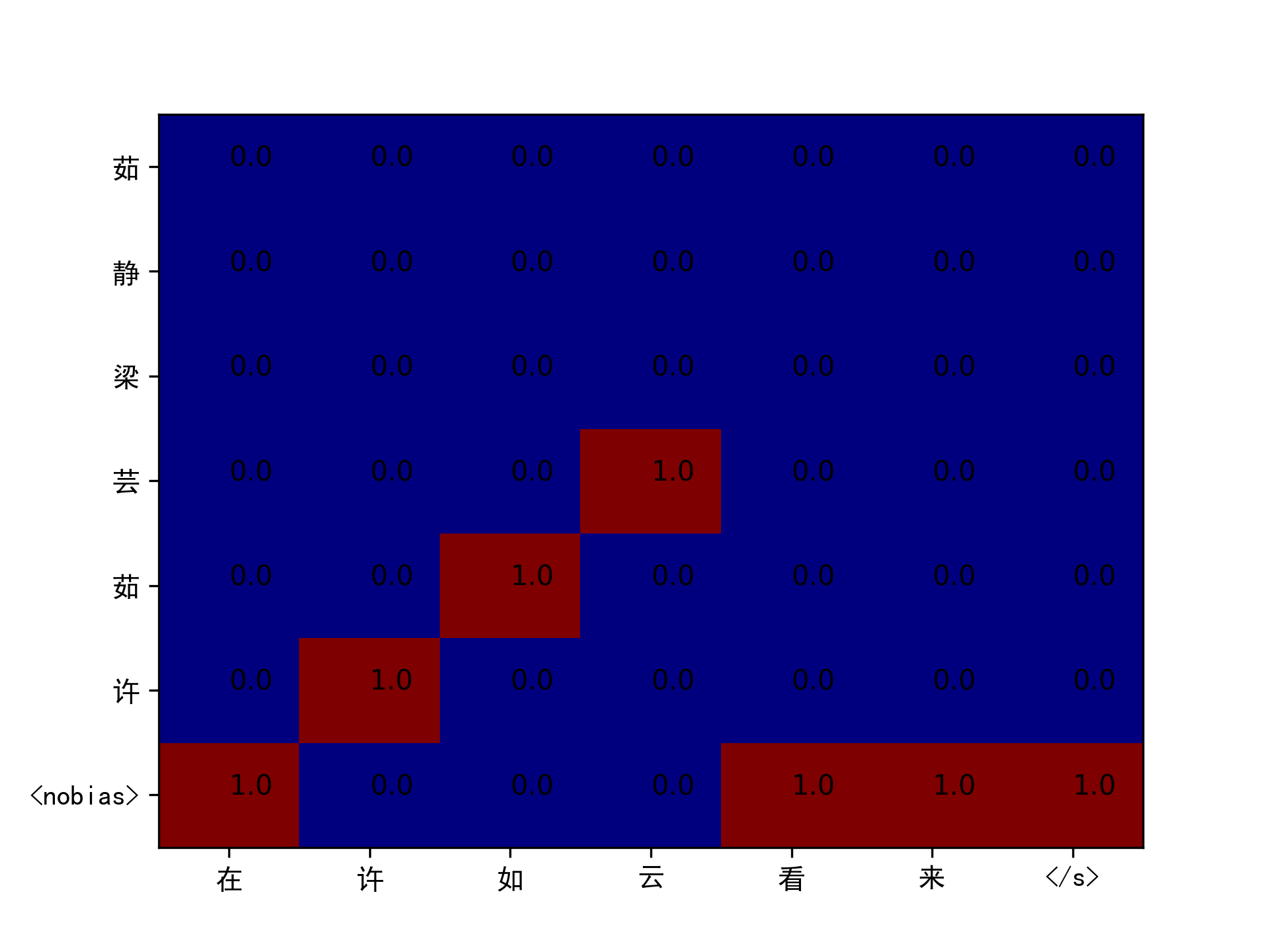,width=6.0cm}}
  \centerline{(f) E7}\medskip
\end{minipage}
\caption{Bias attention map of different models, only one or two bias words were added for simplification.}
\label{fig:att}
\end{figure}

First, we reproduce the CLAS baseline model. However, the recall of the CLAS baseline decreases after adding bias words. We guess it was because too many bias words added during testing (891) compared with training (about 42). So we try to use only the named entities corresponding to each sentence (less than 5) as the bias words list instead of using all named entities. The results are shown in Table \ref{tab:base}. F1-score does not change after adding bias words either, indicating that the number of bias words has nothing to do with bias failure. We then plot the bias attention map of one audio in Fig. \ref{fig:att}(a). The vertical axis represents bias words, and the horizontal axis represents the decoding result. The numbers on the graph represent the attention score. The label is “在许茹芸看来”. So the model should pay attention to the bias word when decoding “许茹芸”. However, it can be seen from the figure that the model only pays attention to the bias word when decoding “许” and ignores it when decoding other words, making “茹芸” recognized as “如云”. Therefore, the CLAS baseline sometimes cannot pay attention to the bias information correctly and it is important to add bias loss to constrain bias attention. The bias loss version in Table \ref{tab:overall} (E1) indicates that recall increased from 78.46\% to 79.59\% when adding bias words. It proves the effectiveness of the bias loss. We also plot the bias attention map of bias loss version in Fig. \ref{fig:att}(b). Although the recognition result is still wrong, the model can basically focus on the correct bias word at the right time. This is the foundation for subsequent optimization.

\subsection{The query of bias attention}
Then, we verify the effect of adding more information to the query of bias attention. Compared with E1 using $d_{t-1}$ only as the query of bias attention, E2 which use $d_{t-1}+y_{t-1}$ can increase recall from 79.59\% to 79.86\%. Moreover, the recall was further raised to 82.16\% when $c_{t}^{x}$ is used as the query of bias attention, indicating the audio contextual information is important for bias attention. We also plot the bias attention map of E3 in Fig. \ref{fig:att}c. Although the recognition result is still wrong, the bias attention score is much stronger than E2 in Fig. \ref{fig:att}b. It shows that the model has more confidence in bias attention, and the bias attention score is more accurate.

\subsection{The granularity of bias information}
In this section, we conduct fine-grained modeling of bias information and compare the effects of different model structure encoding bias words. Fine-grained model E4 in Table \ref{tab:overall} reaches 82.86\% recall outperforming coarse-grained model E3 with 82.16\% recall. We plot the bias attention map of E4 in Fig. \ref{fig:att}d. It can be seen from the figure that it can pay attention to a character in the bias word rather than the whole word, reducing confusion when biasing. Further, after switching the bias encoder from LSTM to BLSTM or conformer, the recall can be improved consistently, achieving 85.11\% and 86.86\% respectively. We plot the bias attention map of E6 in Fig. \ref{fig:att}e. The attention score is much stronger than E4, indicating that better bias encoding can also improve the confidence of bias attention and thus improve the bias effect.

\subsection{Probability fusion}
Although the model's attention to bias words has dramatically increased after the above improvements, the recognition result in Fig. \ref{fig:att} is still wrong. It indicates that even if the model focuses on bias words, it does not mean that the model will use them for recognition since the model relies more on acoustic and linguistic information than bias information in recognition. Therefore, feeding the bias context vector into the decoder cannot utilize the bias information effectively. Considering that the bias attention score in Fig. \ref{fig:att}e has reached one, we can directly use the bias score to correct the output of the model. We compared different correction methods, including pointer generator and interpolation. Table \ref{tab:overall} shows that the optimal bias coefficient of the pointer generator model (E7) is 0.2 with F1-score 93.95\%, which is significantly higher than E6. Merging probabilities of the same characters in different bias words (E8) has little effect. Probably because even if there are multiple identical characters in the bias list, only the character with a specific context has a high bias score. To verify our conjecture, we add another bias word “梁静茹” which has the same character “茹” as “许茹芸” and plot the bias attention map in Fig. \ref{fig:att}f. We can see that only “茹” in “许茹芸” has a high attention score, while the model does not focus on “茹” in “梁静茹” at all. It shows that although the bias score is calculated for each character separately, the adjacent acoustic context is also taken into account, resulting more accurate bias score. And if we merge probabilities of the same characters in the model output and the bias list (E9), F1-score can be further improved to 94.21\%, resulting 53.49\% relative improvement from the CLAS baseline.

\subsection{Prefix tree}
The longest bias word in our previous experiment has only four characters. However, the number of characters in bias words may be vast in the production environment. To verify the effect of the model on long bias words, we retain named entities in AISHELL-NER containing two to sixteen characters and de-duplicate them to get 1186 named entities as the bias words list. The results are shown in Table \ref{tab:long}. It can be seen from the table that the recall of long bias words (5-16) is only 91.75\% in E9, which is much lower than that of short bias words. But after using the prefix tree, the recall is increased to 94.58\%. It shows that the prefix tree is important for the performance of the CLAS model on the long bias words list.

\begin{table}[h!t]
\centering
\caption{Results of long bias words list.}
\label{tab:long}
\begin{tabular}{lccccc}
   \toprule
   Methods & CER & Length & Recall & Precision & F1 \\
   \midrule
   \multirow{3}{*} {E9} & \multirow{3}{*} {4.40} & 2-16 & 94.01 & 94.22 & 94.11 \\
   \multirow{3}{*} {} & \multirow{3}{*} {} & 2-4 & 94.37 & 94.15 & 94.26 \\
   \multirow{3}{*} {} & \multirow{3}{*} {} & 5-16 & 91.75 & 99.49 & 95.46 \\
   \midrule
   \multirow{3}{*} {E10: E9+prefix tree} & \multirow{3}{*} {4.30} & 2-16 & 94.94 & 94.66 & 94.80 \\
   & \multirow{3}{*} {} & 2-4 & 95.03 & 94.54 & 94.78 \\
   & \multirow{3}{*} {} & 5-16 & 94.58 & 99.50 & 96.98 \\
   \bottomrule
\end{tabular}
\end{table}

\section{Conclusions}

In this work, we introduced deep CLAS, a model that can exploit bias information more effectively than CLAS. We analyzed the shortages of CLAS and made a series of improvements. First, we introduce the bias loss forcing model to focus on contextual information. Second, we use richer information as the query of bias attention. Third, we replace phrase-level encoding with character-level encoding to obtain fine-grained contextual information. Fourth, we explicitly use the bias attention score to utilize contextual information more effectively. Finally, we use the prefix tree to reduce the interference of irrelevant bias words. Experiments on the public mandarin ASR data demonstrate a significant improvement of deep CLAS over CLAS.

\section*{Conflict of Interest}

The authors declare that they have no conflict of interest.

\section*{Biographies}

Mengzhi Wang is currently working at iFLYTEK Research in Hefei. He
received his master’s degree in Software Engineering from the University of Science and Technology of China in 2019. His research interests focus on automatic speech recognition.




\nocite{*}
\bibliographystyle{custom-unsrt}
\bibliography{sample.bib}


\end{CJK}
\end{document}